\documentclass{article}

\usepackage{amsmath}
\usepackage{textcomp}

     \usepackage[final]{neurips_2019}

\usepackage[utf8]{inputenc} 
\usepackage[T1]{fontenc}    
\usepackage{wrapfig}

\usepackage{hyperref}       
\usepackage{url}            
\usepackage{booktabs}       
\usepackage{amsfonts}       
\usepackage{nicefrac}       
\usepackage{microtype}      

\usepackage{graphicx}
\usepackage{algorithm}
\usepackage[noend]{algpseudocode}
\usepackage[colorinlistoftodos,prependcaption,textsize=tiny]{todonotes}

\graphicspath{{./}{./plots/}}

\title{FD-Net with Auxiliary Time Steps: Fast Prediction of PDEs using Hessian-Free Trust-Region Methods}
\author{
Nur Sila Gulgec \\
Lehigh University\\
\texttt{sgulgec@gmail.com} \\
\And
Zheng Shi\\
Lehigh University\\
\texttt{zhs310@lehigh.edu} \\
\And
Neil Deshmukh\\
Morovian Academy, Lehigh University\\
\texttt{neil.nitin.de@gmail.com} \\
\And
 Shamim N. Pakzad\\
 Lehigh University\\
 \texttt{snp208@lehigh.edu} \\
 \And
Martin Takac\\
Lehigh University\\
 \texttt{Takac.MT@gmail.com} \\
}

\date{June 2019}

\begin{document}

\maketitle

\begin{abstract}Discovering the underlying physical behavior of the complex systems is crucial, but less well-understood topic in many engineering disciplines. This study proposes a finite-difference inspired convolutional neural network framework to learn the hidden partial differential equations from the given data and iteratively estimate the future dynamical behavior. The methodology designs the filter sizes such that they mimic the finite difference between the neighboring points. By learning the governing equation, the network predicts the future evolution of the solution by using only a few trainable parameters. In this paper, we provide numerical results to compare the efficiency of the second-order Trust-Region Conjugate Gradient (TRCG) method with the first-order ADAM optimizer.
\end{abstract}

\section{Introduction}
Partial differential equations (PDEs) are widely adopted in engineering fields to explain a variety of phenomena such as heat, diffusion, electrodynamics, fluid dynamics, elasticity, and quantum mechanics. With the rapid development in the sensing and storage capabilities provide engineers to reach more knowledge about these phenomena. The collected massive data from multidimensional systems have the potential to provide better understanding of system dynamics and lead to a discovery of more complex systems. 

Exploiting data to discover physical laws has been recently investigated through several studies. \cite{schmidt2009distilling,bongard2007automated} applied symbolic regression and \cite{rudy2017data,schaeffer2017learning}
proposed sparse regression techniques to explain the nonlinear dynamical systems. 
\cite{raissi2018hidden,raissi2017physics} introduced physics informed neural networks using Gaussian processes. \cite{chen2018neural} demonstrated continuous-depth residual networks and continuous-time latent variable models to train ordinary neural networks. \cite{farimani2017deep} proposed conditional generative adversarial networks and \cite{long2017pde} proposed PDE-Net originated from Wavelet theory. 

This study proposes a finite-difference inspired convolutional neural network framework to learn the hidden partial differential equations from the given data and iteratively estimate the future dynamical behavior with only a few parameters. Additionally, we introduce auxiliary time steps to achieve higher accuracy in the solutions.

While first-order methods have been extensively used in training deep neural networks, they struggle to promise the training efficiency. By only considering first-order information, these methods are sensitive to the settings of hyper-parameters, with difficulty in escaping saddle points, and so on. Hessian-free (second-order) \cite{Martens} methods use curvature information, make more progress every iteration, minimize the amount of works of tuning hyper-parameters, and only require Hessian-vector product. In this paper, ADAM \cite{Adam} and TRCG methods \cite{Nocedal} \cite{steihaug} are used to train the proposed network. The empirical results demonstrate that this particular second-order method is more favorable than ADAM to provide high accuracy results to our engineering application of deep learning.


The rest of the paper is organized as follows. First, motivation of our approach is provided in Section \ref{sec2}; then, the proposed methodology is described in Section \ref{sec3}. In Section \ref{sec4} and \ref{sec5}, the numerical study is introduced and main findings of this study are discussed, respectively.

\section{Motivation} \label{sec2}
Let us consider a partial differential equation of the general form
\begin{equation}\label{eq1}
\mathcal {F}(x,t, u,u_t, u_x, u_{xx},u_{xxx}, \dots) = 0
\end{equation}
 where $ \mathcal{F}$ is the non-linear function of $u$, its partial derivatives in time or space where it is denoted by the subscripts. The objective of the study is to implicitly learn the $\mathcal{F}$ from the given time-series measurements at specific time instances and predict the behavior of the equation for long time sequences. 

For easier interpretation of the approach, the proposed algorithm is explained through the motivation problem. Parabolic evolution equations describe processes that are evolving in time. The heat equation is one of the frequently used examples in physics and mathematics to describe how heat evolves over time in an object \cite{incropera2007fundamentals}. Let $u(x,t)$ denotes the temperature at point $x$ at time $t$. The heat equation has the following form for the 1-D bar of length $L$: 
\begin{equation} \label{eq2}
 \tfrac{\partial u}{\partial t} = \beta \tfrac{\partial^2 u}{\partial x^2},
\end{equation}
where $\beta$ is a constant and called the thermal conductivity of the material. Thermal equation has some boundary conditions. If boundaries are perfectly insulated, the boundary conditions are reduced to,
\begin{equation}\label{eq3}
u(0,t) = 0;  u(\pi,t) = 0.
\end{equation}
The PDE of the heat equation can be solved by using Euler method where x and t are discretized for $0 \leq x \leq X $ and $0 \leq t \leq T $ to find directional derivatives.
\begin{equation}\label{eq4}
u(x,t+\Delta t) \approx u(x,t) + \alpha [u(x+\Delta x, t) -2 u(x,t) + u(x-\Delta x, t)],
\end{equation}
where $\alpha = \beta \frac{ \Delta t}{(\Delta x)^2} $. 
When the individual time steps are too from each other, Euler method fails to provide a good solution. The stability criteria is satisfied only when $ \alpha \leq 0.5$ \cite{olsen2011numerical}.
Additionally, for each prediction step, boundary conditions and $\beta$ values are assumed to be known which is not necessarily true for the real applications. In order to address these challenges, data-driven approach is proposed.

\section{Methodology} \label{sec3}
The proposed approach is inspired by the finite difference approximation. Each directional derivative in $\Delta x$ direction is defined as trainable finite difference filters by size of three (i.e., one parameter for the left neighbor, one for the point itself and one for the right neighbor). The trainable parameters only include weights without any nonlinear activation function and biases. When there is a higher degree of partial difference, multiple sets of learnable weights are considered during training. At the boundary conditions, the filter size of two is adopted since there is only one neighbor. The main benefit of using such a filter is to reduce the number of parameters of the network and to use more natural and interpretable building blocks for the engineering applications.


\begin{wrapfigure}{r}{6cm}
  \centering
  \vskip-5pt
  {\includegraphics[trim={0in 0.2in 0 0.2in},width=6cm]{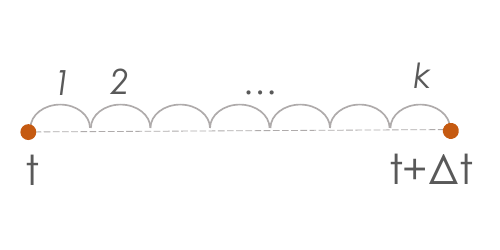}}
  \vskip-5pt
  \caption{Predicting the function with $k$ artificial time steps.}
  \vskip-10pt
  \label{fig1}
\end{wrapfigure}
In order to increase the accuracy and stability, $k$ "artificial" time-steps  are introduced to the network (Figure \ref{fig1}), where $\forall j \in \{0,1,\dots, k\}$ the function value $u(x,t+ \frac{(j+1) \Delta t}{k})$ is computed from the linear combination of the input $u(x,t+j \frac{\Delta t}{k})$ and the feature maps obtained from the difference approximations. These steps are repeated until the prediction of $u(x,t+\Delta t)$. Similar idea is also used in residual neural networks \cite{he2016deep} because of its ease in optimization, however, in our case it is a necessity to obtain solutions for unstable PDEs. The relationship between these iterative updates and Euler discretization is also discussed in the Chen et. al. \cite{chen2018neural}.

{\bf Training:}

Training might take a considerable amount of time while working with long sequences. The proposed approach addresses this problem by training the architecture with randomized mini-batches. We generate samples from randomly picked time intervals during each iteration where $u^n(\cdot, t_i)$ represents a sample from the $n$th time series at time $t = i$. For comparison purposes, first-order ADAM \cite{Adam} and second-order TRCG methods \cite{Nocedal} \cite{steihaug} are used to train the proposed network.



TRCG \cite{Nocedal} method uses Steihaug's Conjugate Gradient (CG)  method \cite{steihaug} to approximately solve the trust region subproblem and obtain a searching direction.  Compared with the exact Newton's method, CG requires only the computation of the Hessian-vector products without explicitly storing the Hessian. This feature makes TRCG method a Hessian-free \cite{Martens} method and suitable to our deep learning application, where the Hessian matrix can be in an immense size due to its quadratic relationship with the number of parameters. To make TRCG more practical to the proposed network and the datasets, a stochastic mini-batch training is adopted: for every iteration of TRCG, one mini-batch dataset is randomly selected to compute the gradient and for CG to compute the Hessian-vector products and solve the trust region subproblem.  

\begin{figure}
  \centering
  \includegraphics[trim={0in 0in 0 0in},width=4.5in]{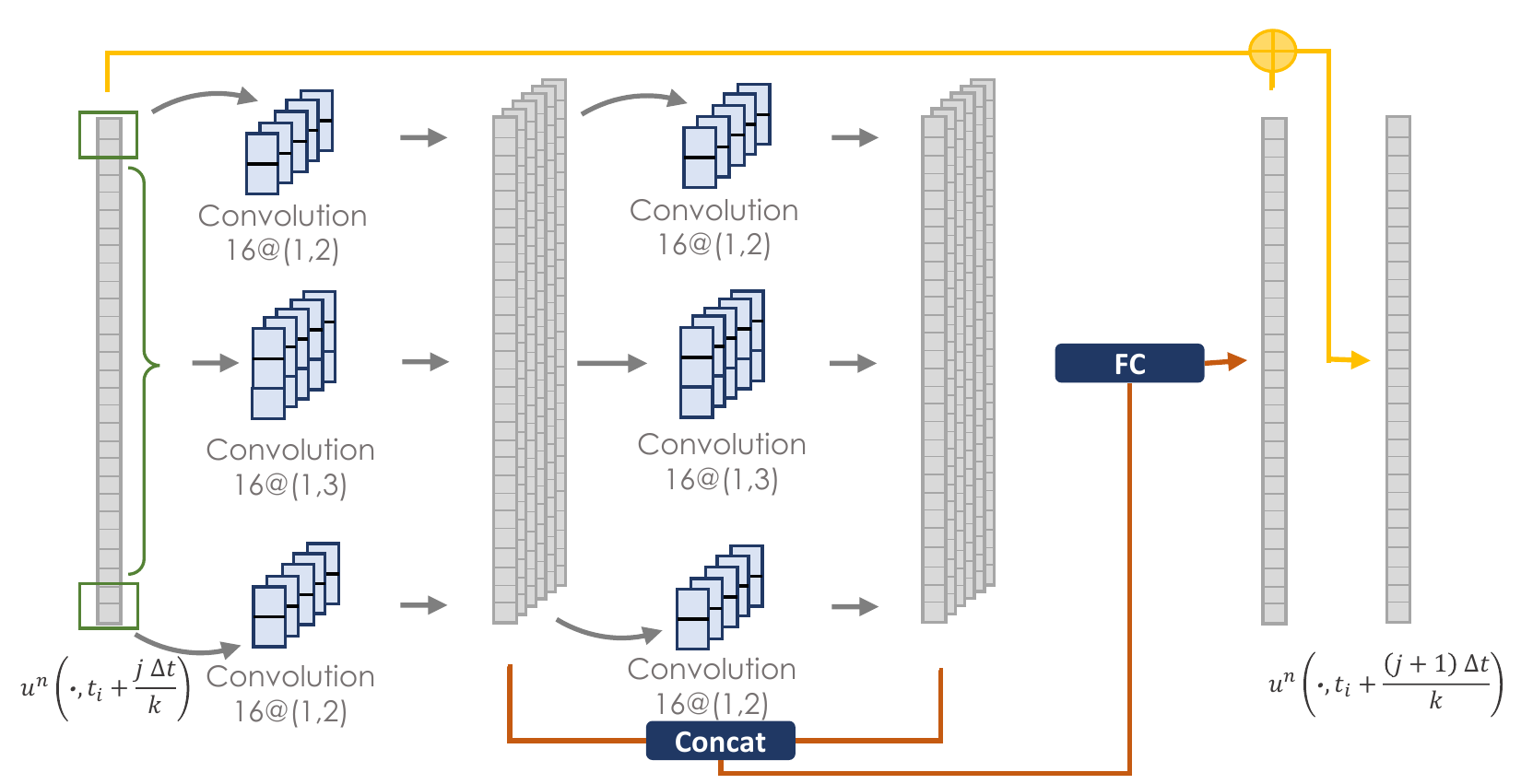}
  \caption{FD-Net predicting the auxiliary time step of $j+1$ from time step $j$.}
  \label{fig2}
  
\end{figure}

{\bf Architecture:}
The general map of the FD-Net architecture is shown in Figure \ref{fig2}. It shows an example of an artificial time step for a selected time $i$ from the sample $n$ generated from the PDE. The sample $u^n(\cdot, t_i+\frac{j\Delta t}{k})$ is passed through two sets of trainable finite difference layers and the resultant of each layer is aggregated through a fully connected (FC) layer. Then, the output of the FC layer are mapped into a residual building block to predict the function behavior at time $t_i+\frac{(j+1)\Delta t}{k}$. The loss function is defined as mean squared error loss between the predicted and true values of the function value. The loss function is penalized more at the boundaries.


\section{Numerical Study} \label{sec4}

A dataset containing $N = 200$ samples are generated with varying initial conditions by selecting different $C_i$ from normal distribution. The the domain of the samples is $ x \in [0,\pi], t \in [0,1000]$ such that total the dataset contains $200\times 31 \times 1000$ values. The dataset is split randomly into train/test sets following an 75/25 ratio. The samples are produced with the parameters $\beta = 0.0002$, $\Delta x = 0.1$ and varying time discretization for stable ($\Delta t = 1$) and unstable ($\Delta t = 200$) cases. 
The boundary conditions and the initial condition of the problem is defined as in \eqref{eq3} and \eqref{eq5}, respectively.
The optimal solution of the heat conduction problem is adopted from the study \cite{olsen2011numerical} and formulated as following:
\begin{equation}\label{eq5}
u(x,0) = \textstyle{\sum}_{i=1}^N C_i \sin (\tfrac{i \pi x}{L}),
\qquad
u(x,t) = \textstyle{\sum}_{i=1}^N C_i \sin (\tfrac{i \pi x}{L}) e^{-   \beta  (\tfrac{i \pi}{L})^2 t}.  
\end{equation}

\section{Results and Discussion} \label{sec5}

During testing, the function value at time $t_{1}$ is predicted by using the function value at time $t_0$. Then, the function value at time $t_{2}$ is predicted by using the function value at time $t_1$. These predictions are repeated for the full length of the sequence. The RMSE of the true and predicted sequence is computed for all $x$'s.

\begin{wrapfigure}{r}{6cm}
\vskip-20pt
    \centering
    {
    \includegraphics[width=6cm,trim={0.5cm 0 0.2cm 0},clip]{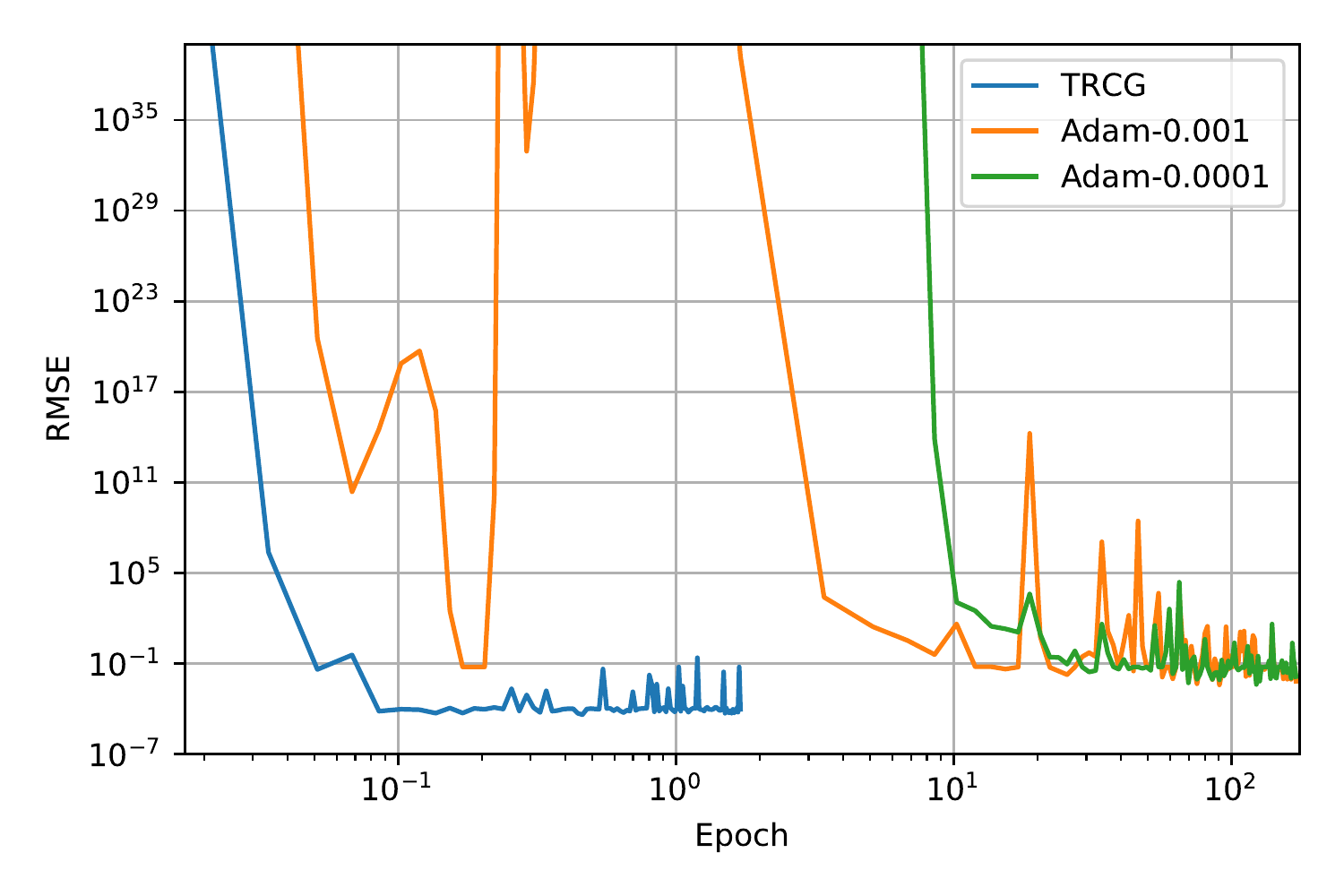}}
    \vskip-10pt 
    \caption{Testing error of FD-Net with $k = 10$.}
    \label{fig3}
    \vskip-20pt 
\end{wrapfigure}
To compare the performance of ADAM and TRCG on training the proposed networks, we conduct experiments with various random seeds and mini-batch sizes on the dataset of the stable case. For Adam, we use two learning rates, $1\mathrm{e}{-3}$ and $1\mathrm{e}{-4}$. For each experiment, depending on the mini-batch size, while we allow ADAM to run between $50$ to $200$ epochs, TRCG is given a small budget, less than $3$ epochs. 

\begin{wrapfigure}{r}{10.2cm}

    \centering
    \vskip-15pt
    \includegraphics[trim={0.4cm 0.4cm 0.4cm 0.4cm},clip,width=5cm]{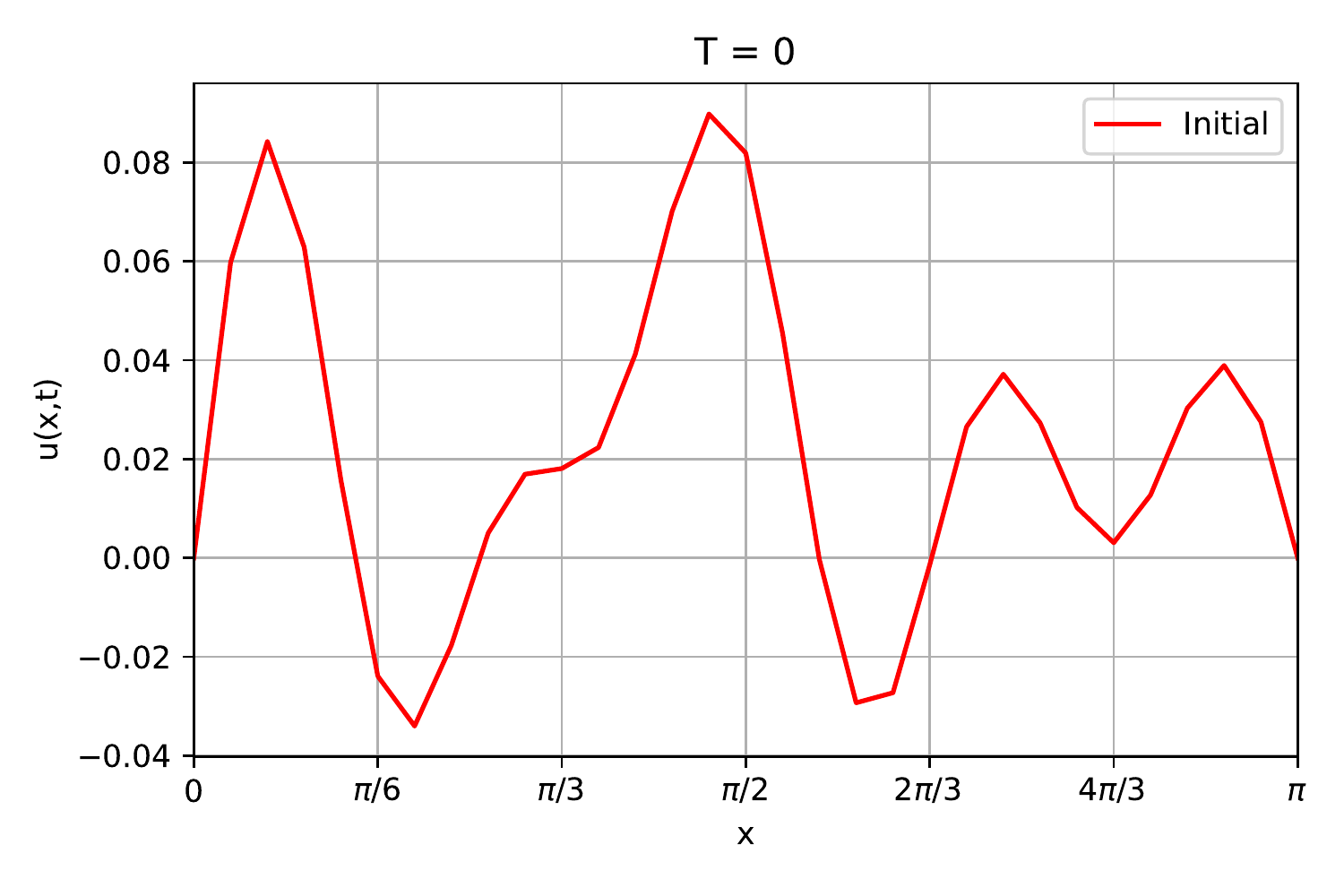}
    \includegraphics[trim={0.4cm 0.4cm 0.4cm 0.4cm},clip,width=5cm]{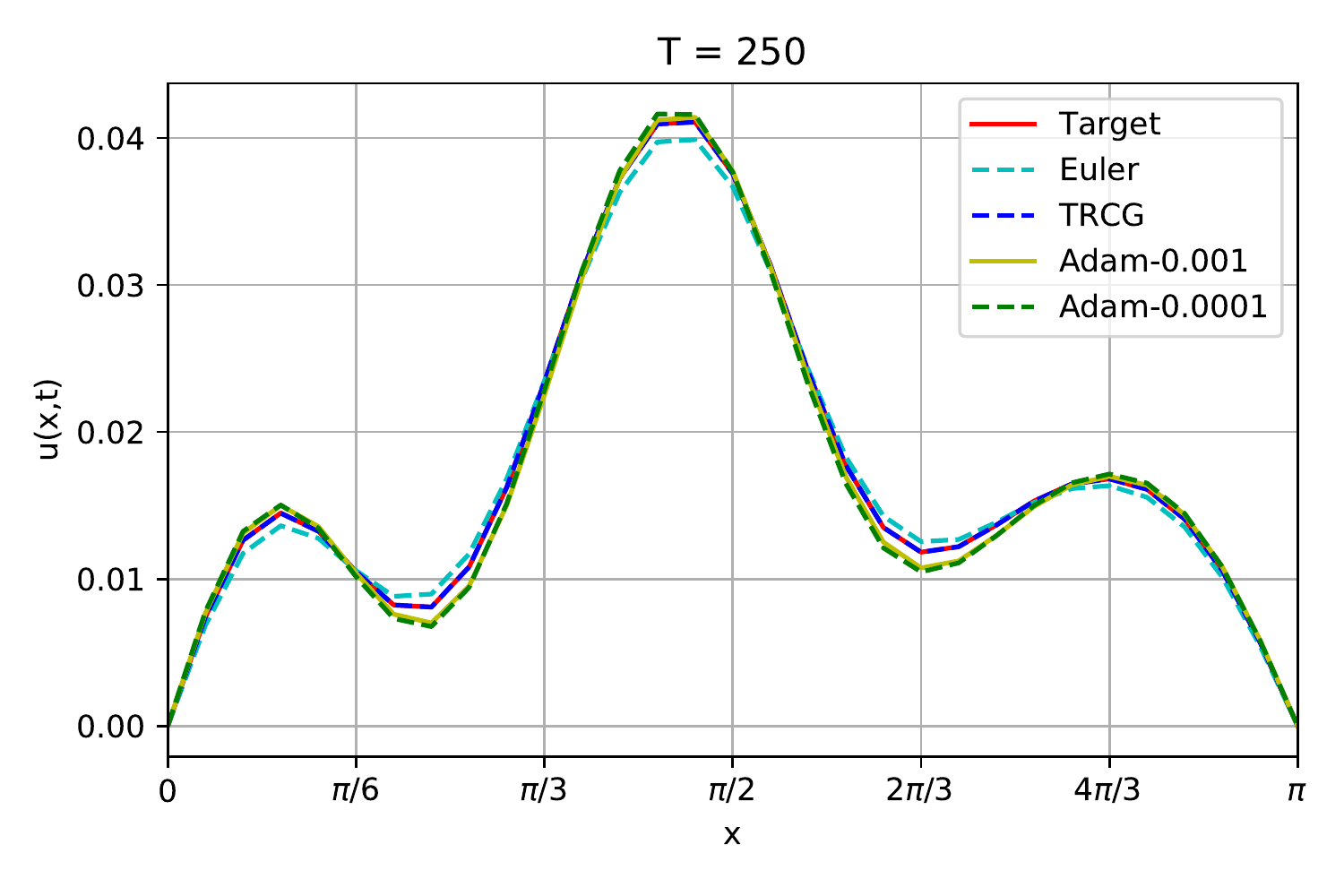}

\vskip-5pt    

    \includegraphics[trim={0.4cm 0.4cm 0.4cm 0.4cm},clip,width=5cm]{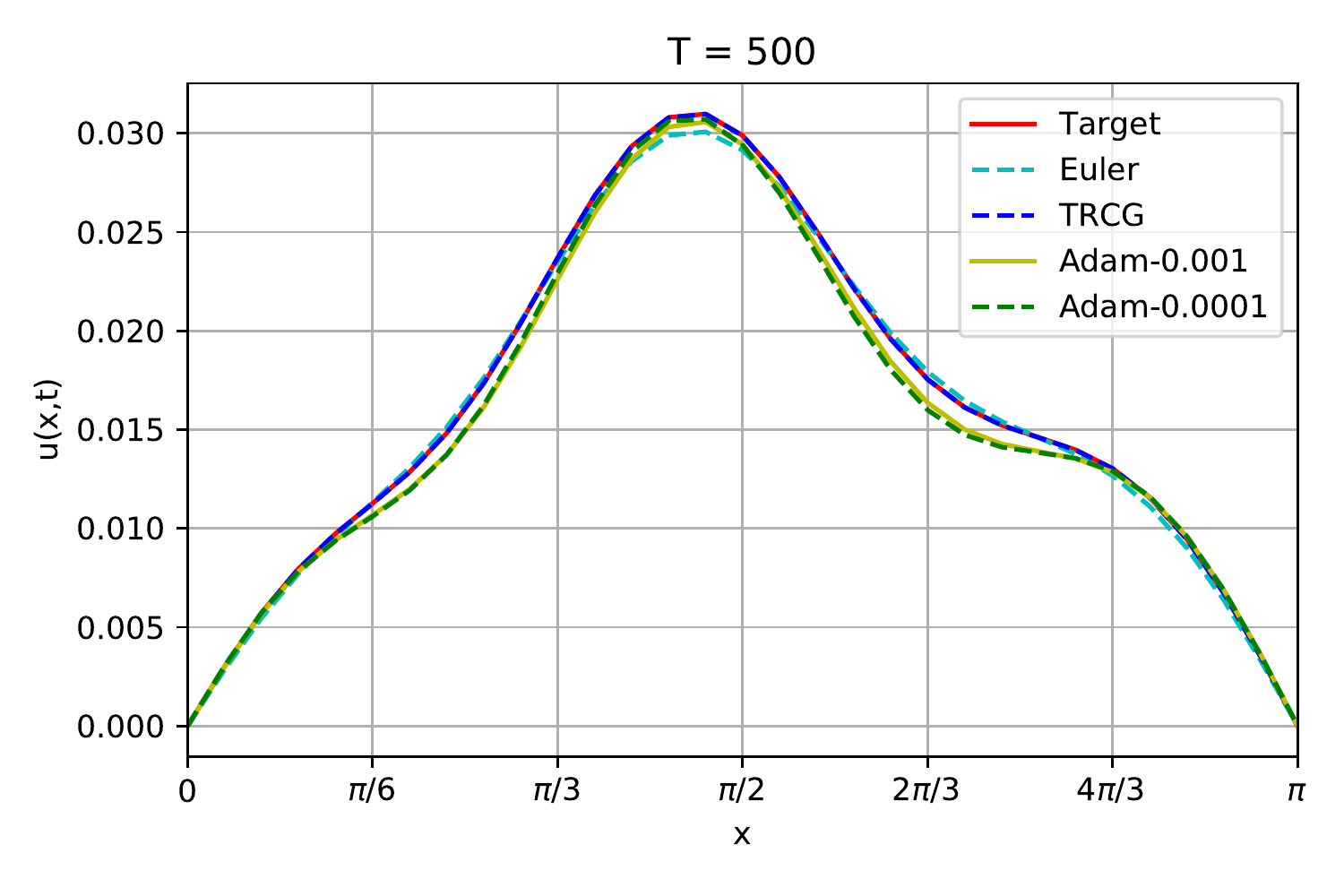}
    \includegraphics[trim={0.4cm 0.4cm 0.4cm 0.4cm},clip,width=5cm]{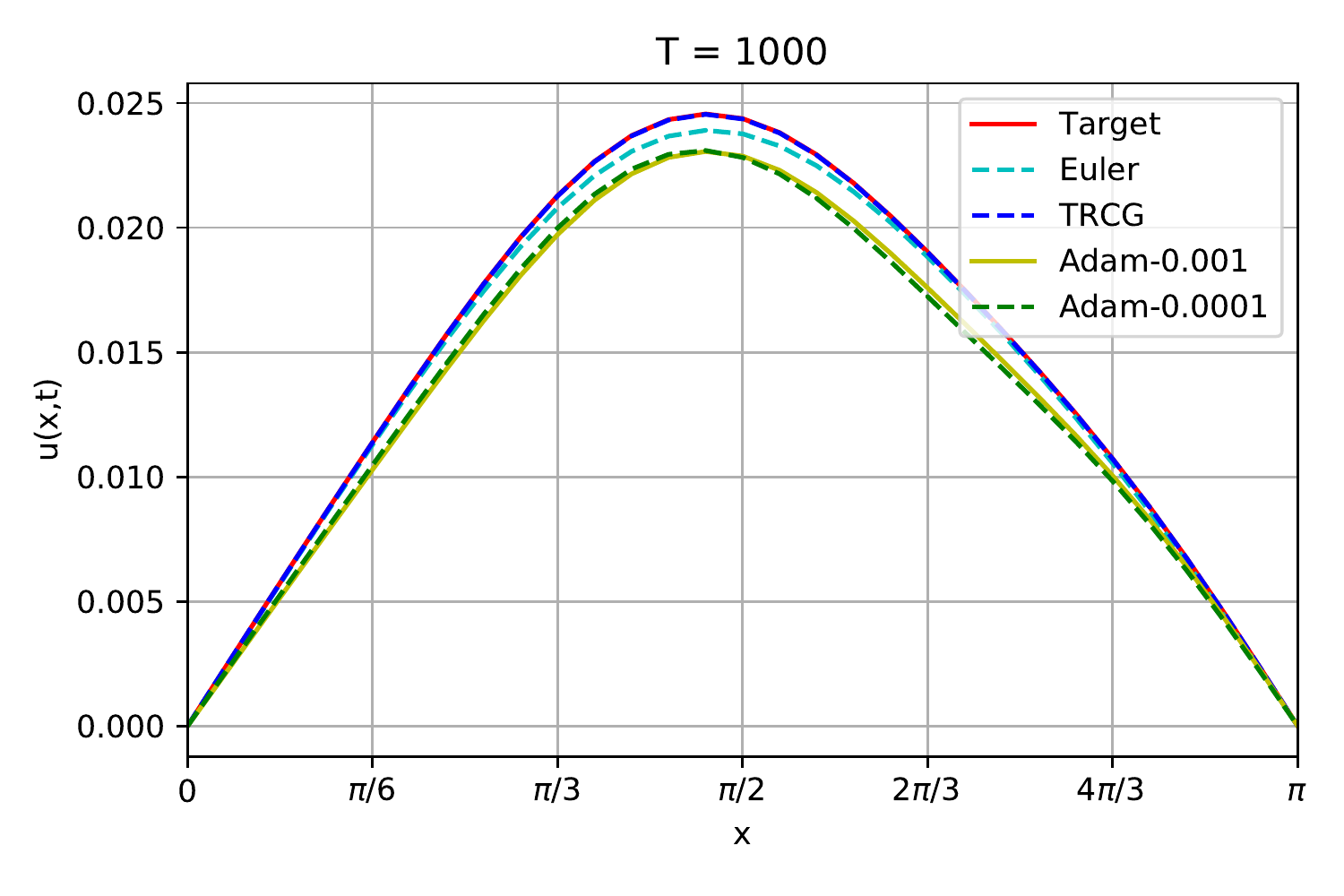}
    \vskip-10pt
    \caption{Prediction of a sample at $ t = 0$, $t=250$, $t=500$, and $t=1000$.}
    \vskip-10pt
    \label{fig4}
\end{wrapfigure}
In spite of the small budget TRCG had, the scale of the testing error in terms of RMSE it achieves at $10^{-5}$, and ADAM is only able to reduce the error to the scale of $10^{-2}$. Figure \ref{fig3} presents an example result from the experiment with the random seed and mini-batch size chosen to be $46$ and $64$, and it illustrates the empirical performance of ADAM and TRCG on the proposed network very well. The results demonstrate a relatively slow convergence of ADAM and suggest that, for the proposed network, second-order information is important and the searching directions that TRCG generated seem to capture the information.   

 The predictions of the testing data is investigated. Figure \ref{fig4} shows the predictions obtained by the proposed method with TRCG and ADAM, and Euler approaches for the time $t = 0$, $t = 250$, $t = 500$ and $t = 1000$. As can be seen from the figure, although the function characteristics change drastically in the longer term, the proposed architecture is able to determine the behavior with only a few parameters. The most accurate prediction is achieved when FD-Net with TRCG method.  
 
 


\begin{wraptable}{r}{8cm}
\vskip-20pt
\centering 
\caption{RMSE for the unstable case.}
\label{table1}
\begin{tabular}{lllll}
\toprule
 & \multicolumn{3}{c}{FD-Net with TRCG} &     \\
 \cmidrule{2-4}
 {\it Batch Size} & {\it k = 1}  & {\it k = 10} & {\it k = 20}  &  {\it Euler} \\
\cmidrule{1-5}
 32     &  0.0345  & 0.0037 & {\bf 0.0028 }  & 73.787 \\ 
 64     &  0.0342  & 0.0038 & {\bf 0.0033 }  & 73.787 \\ 
 128    &  0.0337  & 0.0079 & {\bf 0.0079 }  &  73.787 \\ 
 \bottomrule
\end{tabular}
\vskip-10pt
\end{wraptable} 
 Since the prediction at time instance affects the next time prediction, the effect of the error accumulation is tested for the unstable case with different artificial time steps  $(k = 1)$, $(k = 10)$ and $(k=20)$. Table \ref{table1} shows that the performance of the adopted approach is greater than the Euler approach for the unstable cases and increasing the number of artificial time step increases the accuracy of the method. Although our approach mimics the Euler method when $k=1$, thus better performance is observed.


\section{Acknowledgements}
 Research funding is partially provided by the National Science Foundation through Grant No. CMMI-1351537 by Hazard Mitigation and Structural Engineering program, and by a grant from the Commonwealth of Pennsylvania, Department of Community and Economic Development, through the Pennsylvania Infrastructure Technology Alliance (PITA).
Martin Tak\'a\v{c} was supported by National Science Foundation grants CCF-1618717, CMMI-1663256 and CCF-1740796. 

\bibliographystyle{unsrt}
\bibliography{references}
\end{document}